\documentclass[10pt,letterpaper,journal]{IEEEtran}
\usepackage[utf8]{inputenc}
\usepackage[T1]{fontenc}
\usepackage{tgtermes}
\usepackage{amsmath,amsfonts,amssymb}
\usepackage{array,booktabs,multirow}
\usepackage{graphicx}
\usepackage{capt-of}
\usepackage{cite}
\usepackage{stfloats}
\usepackage{needspace}
\usepackage[hidelinks]{hyperref}
\begin{document}
\title{ARC-VLN: Adaptive Recursive Computing-based Visual-Language Navigation}
\author{Xiang~Zhang, Xiangyu~Dong, Yang~Lu, and Xiaoguang~Ma%
\thanks{This work was supported in part by the Hunan Key Scientific and
Technological Research Foundation under Grant 2024QK2001.
Xiang Zhang and Xiangyu Dong contributed equally to this work.
(Corresponding authors: Yang Lu and Xiaoguang Ma.)}%
\thanks{Xiang Zhang is with the School of Artificial Intelligence,
Henan University, Zhengzhou 450046, China, and also with the School of
Computer and Information Engineering, Henan University, Kaifeng 475001,
China (e-mail: \href{mailto:Z408139588@163.com}{Z408139588@163.com}).}%
\thanks{Xiangyu Dong is with the College of Information Science and
Engineering, Northeastern University, Shenyang 110819, China
(e-mail: \href{mailto:dxy1999ai@163.com}{dxy1999ai@163.com}).}%
\thanks{Yang Lu is with the School of Artificial Intelligence,
Henan University, Zhengzhou 450046, China
(e-mail: \href{mailto:lykyc@henu.edu.cn}{lykyc@henu.edu.cn}).}%
\thanks{Xiaoguang Ma is with the School of Artificial Intelligence, Tianjin University,
Tianjin 300350, China (e-mail: \href{mailto:coltlight@163.com}{coltlight@163.com}).}%
}
\maketitle
\begin{abstract}
Efficient aerial visual-language navigation (VLN) requires reducing model storage and inference costs while maintaining instruction-conditioned action prediction. We present ARC-VLN, a token-type-aware recursive computation framework built on a language backbone with shared parameters. ARC-VLN retains full-depth computation for text and action tokens while adaptively selecting exit depths for visual tokens. Representations of early-exiting visual tokens are retained, and active-token compaction shortens the sequences processed by subsequent recursive blocks. To mitigate routing collapse to a single exit depth, we design a dedicated learning strategy that encourages the model to distribute visual tokens across multiple exit depths. On OpenFly’s Test-Seen and Test-Unseen subsets, ARC-VLN achieves the lowest navigation error and the highest success rate and success weighted by path length among all compared methods. Meanwhile, it reduces the average number of decision steps per trajectory, total floating-point operations, mean model inference time per trajectory, and average GPU memory usage during inference, indicating its great potential as a VLN backbone for lightweight applications.
\end{abstract}
\begin{IEEEkeywords}
Efficient inference, recursive parameter sharing, token-type-aware routing, visual-language navigation.
\end{IEEEkeywords}
\section{Introduction}

Visual-language navigation (VLN) combines language understanding and visual perception to guide embodied agents according to natural language instructions~\cite{zhang2025flexvln,tan2025sourcefree}. In recent years, VLN studies such as CityNav~\cite{lee2025citynav}, OpenUAV~\cite{wang2025openuav} and OpenFly~\cite{gao2026openfly} have tightly integrated vision-language-action (VLA) models with perception and path-planning modules for aerial VLN. This integration enables natural language instruction parsing, semantic understanding of unfamiliar environments and autonomous navigation without prior maps, substantially improving aerial navigation in complex environments. To further improve VLN performance, studies such as FlightGPT~\cite{cai2025flightgpt} and FreeFly-Thinking~\cite{zhou2026freeflythinking} have introduced chain-of-thought (CoT) reasoning. By decomposing navigation tasks and modeling spatial relationships through multi-step logical reasoning, these methods strengthen long-horizon planning and autonomous decision-making in complex environments.

As large-scale generative VLA models are adopted for aerial VLN, parameter counts and the computational cost of inference increasingly constrain practical deployment. On the one hand, navigation architectures built around large language models (LLMs) contain large numbers of unique parameters, imposing substantial model storage requirements. On the other hand, tokens in multimodal sequences serve distinct task roles, so uniformly processing all tokens through the full model depth may incur redundant computation.

To address these challenges, we present ARC-VLN, an efficient aerial VLN framework based on adaptive recursive computing. The framework reuses weights across recursive stages of the LLM backbone to reduce the number of unique parameters. Building on this shared backbone, we introduce token-type-aware routing that allocates computational resources according to the roles of text, visual and action tokens in navigation.

Our main contributions are as follows:\par\nopagebreak[4]

(1) We propose a token-type-aware adaptive recursive computation framework for aerial VLN, which retains full recursive depth for text and action tokens while adaptively assigning exit depths to visual tokens. The sequences processed by subsequent recursive blocks are shortened through active-token compaction, reducing computation within the shared backbone.

(2) We incorporate cross-entropy regularization against a uniform target into the auxiliary routing loss alongside load-balancing loss and z-loss, providing a non-zero corrective gradient to the corresponding logit even as the routing probability of an exit depth approaches zero. This regularization mitigates routing collapse to a single exit depth and promotes visual-token assignments across multiple exit depths.

(3) On the evaluated OpenFly Test-Seen and Test-Unseen subsets, ARC-VLN uses approximately 2.7B unique language-backbone parameters and achieves the lowest navigation error and the highest success rate and success weighted by path length among the compared methods, together with greater computational efficiency.

\section{Related Work}

\subsection{Visual-Language Navigation}

VLN connects language instructions with visual observations to guide navigation. Early work, such as the R2R benchmark introduced by Anderson et al.~\cite{anderson2018vln}, first integrated vision–language models with embodied navigation, accelerating the development of VLN models. Subsequent studies, such as REVERIE~\cite{qi2020reverie} and SOON~\cite{zhu2021soon}, further expanded the range of semantic granularity in instructions and increased target diversity. However, the topological assumptions of discrete indoor navigation graphs fundamentally limit the transferability of these approaches to unconstrained real-world environments.

Aerial VLN benchmarks extend this setting towards outdoor scenes and more realistic flight. AerialVLN~\cite{liu2023aerialvln} establishes language-guided trajectory planning in continuous three-dimensional space, while OpenUAV~\cite{wang2025openuav} expands task coverage through aerial object search with varying levels of navigation assistance. CityNav~\cite{lee2025citynav} and OpenFly~\cite{gao2026openfly} further broaden semantic and environmental coverage through geography-aware instructions and greater scene diversity, respectively. To further improve understanding of complex scenes and long-horizon reasoning in VLN models, CoT-based methods~\cite{lin2026evolvenav}, including FlightGPT~\cite{cai2025flightgpt} and FreeFly-Thinking~\cite{zhou2026freeflythinking}, connect environmental perception with action generation through explicit intermediate reasoning.

These advances have improved VLN performance and expanded its reasoning capability. However, reliance on large models introduces substantial storage demands and inference costs, posing challenges for real-time aerial VLN on resource-constrained platforms.

\subsection{Large Model Compression and Efficient Inference}

Parameter sharing reduces model storage by reusing weights across computational depth. Cross-layer sharing methods, exemplified by ALBERT~\cite{lan2020albert} and MobileLLM~\cite{liu2024mobilellm}, tie weights across layers or adjacent blocks to reduce unique parameters. Among recursive language models, Relaxed Recursive Transformers~\cite{bae2025relaxed} combine cross-layer parameter sharing with layer-wise low-rank adaptation to reduce parameter count while mitigating performance degradation. MoR~\cite{bae2025mor} combines parameter sharing with adaptive recursion for text tokens, improving both parameter and computational efficiency in language modeling.

Dynamic computation reduces inference costs by allocating computational resources according to input requirements. Sequence-level early-exit methods, such as DeeBERT~\cite{xin2020deebert} and FastBERT~\cite{liu2020fastbert}, terminate inference based on prediction confidence but cannot independently assign processing depths to tokens within the same input. Token-level routing methods, including Mixture of Depths (MoD)~\cite{raposo2024mod} and p-MoD~\cite{zhang2025pmod}, provide finer-grained control by selecting which tokens participate in each layer’s computation. In addition, Mixture of Experts (MoE) controls computation through sparse expert activation. Across language models~\cite{jiang2024mixtral,dai2024deepseekmoe} and multimodal models~\cite{lin2026moellava}, this approach expands model capacity while activating only a subset of experts for each token. These methods improve computational efficiency through selective execution, but do not inherently reduce parameter storage, highlighting the need to address both costs in resource-constrained aerial VLN. To address storage and computational constraints in aerial VLN, we combine recursive parameter sharing with token-type-aware routing, retaining full recursive depth for instruction and action tokens while adaptively assigning computational depths to visual tokens.

\section{Methods}

\subsection{Problem definition}

Aerial VLN requires the model to predict the navigation action for the current state based on natural language instructions and continuous visual observations. Let $\mathcal{T}$ denote text tokens, $\mathcal{V}$ visual tokens obtained by encoding visual observations with a visual encoder, and $\mathcal{A}$ action tokens. The aerial VLN model can then be formulated as:

\begin{equation}
\mathcal{A}_t=f_{\theta}(\mathcal{T},\mathcal{V},h_{t-1}),
\label{eq:29}
\end{equation}
where $\mathcal{A}_t$ denotes the navigation action predicted at time step $t$, $h_{t-1}$ represents the historical navigation state, and $\theta$ denotes the model parameters.

We investigate the trade-offs among navigation performance, unique parameter count and inference cost under this formulation.

\subsection{ARC-VLN framework}

\begin{figure*}[!t]
\centering
\includegraphics[width=1.00\textwidth]{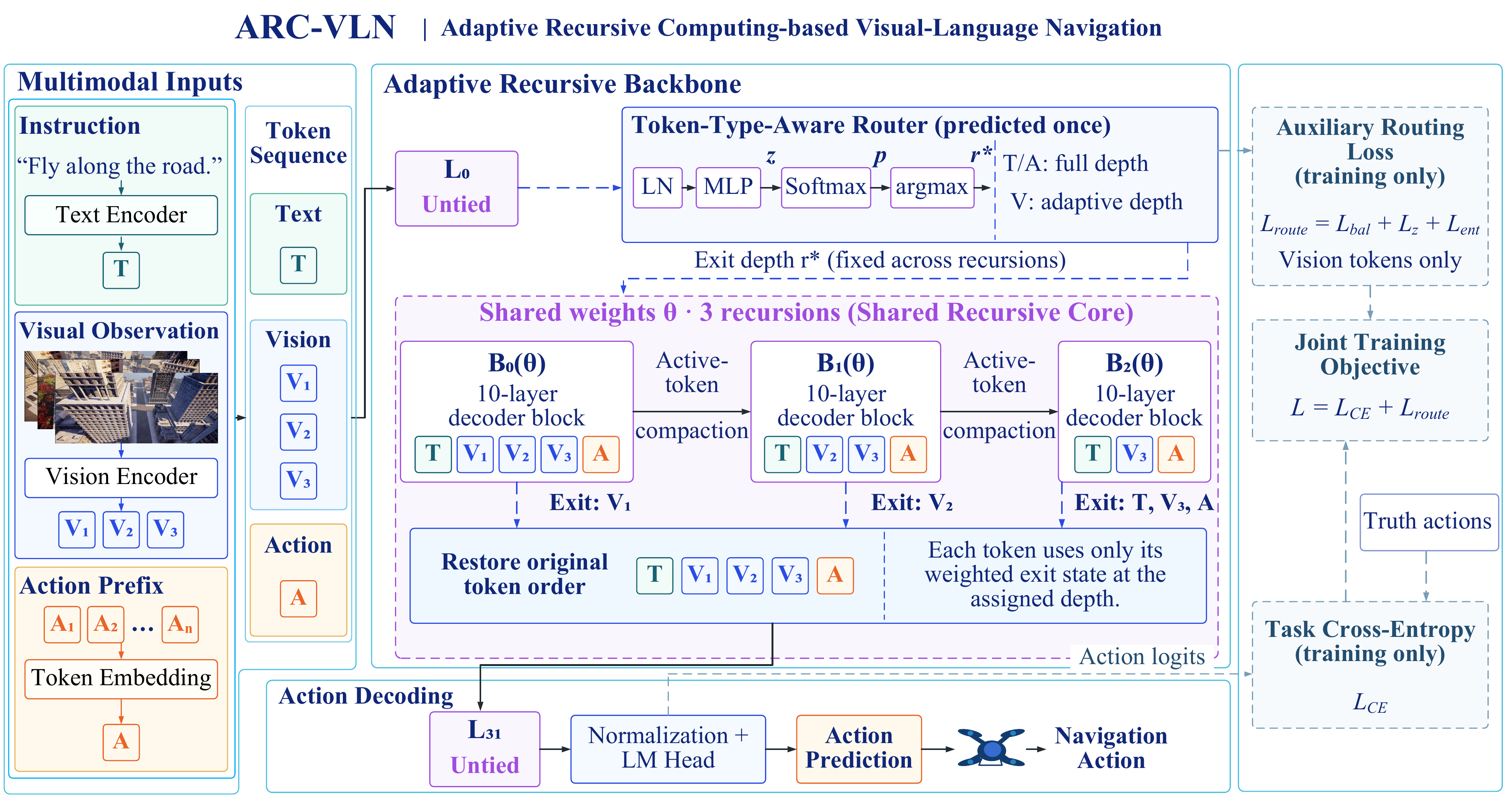}
\caption{Overview of ARC-VLN with recursive parameter sharing and token-type-aware routing.}
\label{fig:1}
\end{figure*}

Fig.~\ref{fig:1} shows ARC-VLN, which tokenizes and embeds instructions into text representations. A visual encoder maps multiframe observations into the LLM hidden space. The input sequence comprises text tokens, visual tokens and an action prefix, in that order. Under causal attention, visual tokens can thus attend to preceding instructions.

After an initial unshared Transformer layer, the sequence enters recursive blocks that reuse intermediate-layer weights. Before recursion, a router assigns visual tokens exit depths, while text and action tokens retain the full depth. Active-token compaction excludes exited visual tokens from subsequent blocks, shortening their input sequences.

After recursive processing, all token representations are assembled and passed through a final unshared Transformer layer. The language model head then produces a probability distribution over action tokens, which is decoded into the navigation action for the current time step. The unmanned aerial vehicle (UAV) executes the action and acquires new visual observations for the next navigation decision.

\subsection{Recursive blocks with parameter sharing}

In an $N$-layer Transformer decoder, the first and last layers remain unshared. The $N - 2$ intermediate layers form $R$ recursive blocks, each containing $B = (N - 2) / R$ layers with distinct parameter sets. For $i = 0, 1, \ldots, B - 1$, corresponding layers across blocks share the same \texttt{nn.Parameter} objects through hard parameter tying. Each shared parameter set is initialized by averaging the corresponding pretrained weights. Parameter sharing reduces unique parameters while preserving the maximum effective computational depth.

For the 32-layer LLaMA-2-7B model, setting $R=3$ retains the first and last layers as unshared layers. The 30 intermediate layers are reorganized into a block of 10 layers with distinct parameter sets, executed three times. This preserves an effective computational depth of 32 layers while reducing the number of unique decoder layers to 12. The number of unique parameters in the LLM backbone consequently decreases from approximately 6.7B to approximately 2.7B.

\subsection{Token-type-aware routing}

Text tokens encode navigation goals and action constraints, while action tokens contribute to control sequence generation; both retain full recursive depth. Visual exit depths are predicted from pre-recursion hidden states, allowing token-specific computation within each observation. Exited tokens retain their outputs, while active tokens continue through subsequent recursive blocks. Outputs are restored to their original order and passed through the final unshared Transformer layer. Normalization and the language model head then produce action-token predictions.

\subsubsection{Router architecture}

The router uses a two-layer multilayer perceptron (MLP) to predict routing scores. Its input is the hidden state $\boldsymbol{h}\in\mathbb{R}^{d}$ produced by the initial unshared Transformer decoder layer (layer 0). The router first applies layer normalization to remove scale differences, then computes routing scores through two linear transformations:

\begin{equation}
\hat{\boldsymbol{h}}=\operatorname{LayerNorm}(\boldsymbol{h}),
\label{eq:45}
\end{equation}

\begin{equation}
\boldsymbol{z}=\boldsymbol{W}_2\sigma\!\left(\boldsymbol{W}_1\hat{\boldsymbol{h}}\right),
\label{eq:46}
\end{equation}
where $\mathbb{R}^{d}$ denotes the $d$-dimensional real vector space. The matrices $\boldsymbol{W}_1\in\mathbb{R}^{2d\times d}$ and $\boldsymbol{W}_2\in\mathbb{R}^{R\times 2d}$ contain the weights of the first and second linear transformations, respectively. The activation function is denoted by $\sigma(\cdot)$, and $\boldsymbol{z}$ contains the routing scores for the recursive depths.

Applying softmax to $\boldsymbol{z}$ yields the routing probability for each recursive depth:

\begin{equation}
p_r=\frac{\exp(z_r)}{\sum_{j=0}^{R-1}\exp(z_j)},\quad r=0,1,\ldots,R-1,
\label{eq:50}
\end{equation}
where $p_r$ is the probability of assigning the token to exit after recursive block $r$. The router selects the depth with the highest routing probability:

\begin{equation}
r^*=\operatorname{arg\,max}\;p_r.
\label{eq:52}
\end{equation}

The target exit depth $r^*$ indicates that the token is processed sequentially by recursive blocks $0,1,\ldots,r^*$ before exiting.

Given variable visual information density, the final rule assigns adaptive depths to visual tokens while retaining full depth for text and action tokens:

\begin{equation}
r^*_{\mathrm{final}}=\begin{cases}R-1,&\text{if }\tau=0,2\\r^*,&\text{if }\tau=1\end{cases},
\label{eq:55}
\end{equation}
where $\{\tau=0,1,2\}$ denotes text, visual and action tokens, respectively. Once routing decisions are fixed, computation proceeds sequentially through recursive blocks indexed by $r=0,1,\ldots,R-1$.

\subsubsection{Adaptive-depth execution}

The active-token set for recursive block $r$ is defined as

\begin{equation}
\mathcal{V}_r=\left\{t\mid r^*_{\mathrm{final}}(t)\ge r\right\},
\label{eq:59}
\end{equation}
which contains all tokens that require processing by recursive block $r$. For the first recursive block ($r=0$), $\mathcal{V}_0$ contains all tokens, and the full sequence is processed. For subsequent recursive blocks ($r\ge1$), we use a Gather-Compute-Scatter strategy to avoid redundant computation on inactive tokens.

(1) \textbf{Gather: }The hidden states of active tokens in $\mathcal{V}_r$ are gathered from the full sequence and compacted into a subsequence. The corresponding causal attention mask and rotary positional encodings are gathered alongside them to preserve the original positional information.

(2) \textbf{Compute: }The compacted subsequence is processed by the $B$ Transformer decoder layers in recursive block $r$.

(3) \textbf{Scatter: }The computed hidden states are written back to their corresponding positions in the full sequence. After each recursive block, tokens satisfying ($r^*_{\mathrm{final}}=r$) have reached their target exit depths. Their weighted outputs are incorporated into the final representations as follows:

\begin{equation}
\boldsymbol{o}_t=\boldsymbol{h}^{(0)}_t+\boldsymbol{h}^{(r^*)}_t\cdot w_t.
\label{eq:64}
\end{equation}

Here, $\boldsymbol{h}^{(0)}_t$ denotes the original hidden state of token $t$ before entering the first recursive block. The term $\boldsymbol{h}^{(r^*)}_t$ denotes the output of token $t$ at its target recursive block, and $w_t$ is the corresponding routing weight.

\subsection{Auxiliary routing loss}

The router weights $\boldsymbol{W}_1$ and $\boldsymbol{W}_2$ are learned end to end through two gradient pathways. The main-task cross-entropy loss supplies gradients indirectly through the differentiable routing weight $w_t$. A dedicated auxiliary routing loss provides direct training signals to improve training stability and mitigate routing collapse. This auxiliary loss is computed only for visual tokens and comprises three components.

\textbf{(1)} \textbf{Load-balancing loss}\par\nopagebreak[4]

To discourage assigning all visual tokens to a single depth, we use a load-balancing loss. This loss sums the products of mean routing probabilities and actual assignment frequencies:

\begin{equation}
\begin{aligned}
\mathcal{L}_{\mathrm{bal}} &= \alpha\cdot R\cdot\sum_{r=0}^{R-1}P_r\cdot f_r,\\
P_r &= \frac{1}{|\mathcal{V}_c|}\sum_{t\in\mathcal{V}_c}p_r^{(t)},\\
f_r &= \frac{|\{t\in\mathcal{V}_c\mid r^*(t)=r\}|}{|\mathcal{V}_c|}.
\end{aligned}
\label{eq:70}
\end{equation}

Here, $P_r$ denotes the mean routing probability for depth $r$ across all visual tokens, and $f_r$ is the fraction of visual tokens assigned to that depth. The set $\mathcal{V}_c$ contains the visual tokens in the current batch, and $\alpha$ is the load-balancing coefficient.

By jointly accounting for routing probabilities and assignment frequencies, this loss penalizes concentration at a few recursive depths, thereby mitigating load imbalance across exit depths.

\textbf{(2) Z-loss}\par\nopagebreak[4]

To improve numerical stability during router training, we introduce z-loss to regularize logit magnitudes:

\begin{equation}
\mathcal{L}_z=\beta\cdot\frac{1}{|\mathcal{V}_c|}\sum_{t\in\mathcal{V}_c}\left(\log\sum_{r=0}^{R-1}\exp(z_r^{(t)})\right)^2.
\label{eq:75}
\end{equation}

Here, $\beta$ is the coefficient of z-loss. By penalizing the squared logsumexp, this loss limits numerical drift in the log-normalization factor, providing an additional stability constraint during router training.

\textbf{(3)} \textbf{Cross-entropy regularization loss}\par\nopagebreak[4]

With only load-balancing loss and z-loss, we observed visual tokens concentrating at one exit depth. We therefore introduce cross-entropy regularization against a uniform target to strengthen gradients for low-probability depths:

\begin{equation}
\mathcal{L}_{\mathrm{ent}}=\gamma\cdot\frac{1}{|\mathcal{V}_c|}\sum_{t\in\mathcal{V}_c}\left(-\frac{1}{R}\sum_{r=0}^{R-1}\log p_r^{(t)}\right),
\label{eq:79}
\end{equation}
where $\gamma$ represents the cross-entropy regularization coefficient.

The gradient of the regularizer with respect to logit $z_j$ satisfies
\begin{equation}
\setlength{\abovedisplayshortskip}{\abovedisplayskip}
\frac{\partial\mathcal{L}_{\mathrm{ent}}}{\partial z_j}\propto p_j-\frac{1}{R}.
\label{eq:81}
\end{equation}
Thus, when the routing probability of a depth approaches zero, the regularizer still provides a non-zero corrective gradient. This discourages excessive concentration of routing probabilities and reduces the risk of collapse to a single exit depth.

The total auxiliary routing loss is the sum of the three components:

\begin{equation}
\mathcal{L}_{\mathrm{route}}=\mathcal{L}_{\mathrm{bal}}+\mathcal{L}_z+\mathcal{L}_{\mathrm{ent}}.
\label{eq:83}
\end{equation}

The final training loss combines the main-task loss $\mathcal{L}_{\mathrm{CE}}$ with the auxiliary routing loss:

\begin{equation}
\mathcal{L}=\mathcal{L}_{\mathrm{CE}}+\mathcal{L}_{\mathrm{route}}.
\label{eq:85}
\end{equation}

\section{Experiments}

\subsection{Experimental setup}

\subsubsection{Dataset and evaluation subset}

We train and evaluate ARC-VLN on OpenFly~\cite{gao2026openfly}, which contains 100,000 flight trajectories across 18 distinct scenes. These scenes combine simulated environments with point-cloud reconstructions of real-world settings, providing visual diversity and realism that help limit overfitting to one rendering style. We use the full OpenFly training split and evaluate on two test subsets, Test-Seen and Test-Unseen. Test-Seen contains 1,392 trajectories across seven seen scenes, while Test-Unseen contains 404 trajectories from one unseen scene. The evaluation comprises 1,796 trajectories in total.

\subsubsection{Evaluation metrics}

We evaluate ARC-VLN across three dimensions: task performance, parameter efficiency and computational efficiency. Task performance is measured using navigation error (NE), success rate (SR), oracle success rate (OSR) and success weighted by path length (SPL). NE is the mean Euclidean distance between the UAV’s final position and the goal, measured in meters (m). SR is the proportion of episodes ending less than 20 m from the goal. OSR is the proportion of episodes entering the same goal region at any point along the trajectory. SPL weights successful episodes by the ratio of shortest-path length to actual flight distance. Failed episodes contribute zero, and the scores are averaged across all episodes.

Parameter efficiency is measured by the number of unique LLM backbone parameters (Par), reported in billions (B), with shared parameters counted only once. Computational efficiency is evaluated using average steps (AS), floating-point operations (GFL), inference latency (IL) and average inference memory (AIM). AS is the mean number of decision steps per episode, with fewer steps indicating more compact decision sequences when SR is maintained. GFL is the total floating-point operation count for complete inference across all evaluation samples, expressed in GFLOPs. IL is the mean forward-computation time per trajectory, measured in seconds (s), excluding environmental interaction overhead such as simulator image acquisition and pose updates. AIM is the mean GPU memory usage during inference, measured in gigabytes (GB).

\subsubsection{Implementation details}

We use OpenVLA-7B with a LLaMA-2-7B backbone and dual-stream fusion of SigLIP-ViT-SO and DINOv2-ViT-L visual features. We use $R = 3$ recursive blocks, with coefficients $\alpha = 1 \times 10^{-2}$, $\beta = 1 \times 10^{-5}$ and $\gamma = 1 \times 10^{-2}$ for load-balancing loss, z-loss and cross-entropy regularization, respectively. To avoid early representation collapse, the routing warmup value decreases linearly from 1.0 to 0 over the first 3,000 training steps. Training uses three epochs, a global batch size of 16 and a learning rate of $2 \times 10^{-4}$.

For the experiments reported in Tables I and III, each method or model variant is evaluated seven times on each fixed test subset (Test-Seen and Test-Unseen). Evaluation metrics are computed separately for each subset in each run, and the arithmetic mean of each metric across the seven runs is reported.

\begin{table*}[!t]
\caption{Comparison of navigation performance, parameter efficiency and computational efficiency on the OpenFly benchmark}
\label{tab:1}
\centering
\begingroup
\footnotesize
\setlength{\tabcolsep}{3pt}
\renewcommand{\arraystretch}{1.12}
\begin{tabular*}{\textwidth}{@{\hspace{6pt}\extracolsep{\fill}}*{10}{c}@{\hspace{6pt}}}
\toprule
\multicolumn{10}{c}{Test-Seen}\\
\midrule
Method & NE\ensuremath{\downarrow} & SR\ensuremath{\uparrow} & OSR\ensuremath{\uparrow} & SPL\ensuremath{\uparrow} & Par\ensuremath{\downarrow} & AS\ensuremath{\downarrow} & GFL\ensuremath{\downarrow} & IL\ensuremath{\downarrow} & AIM\ensuremath{\downarrow}\\
\midrule
AerialVLN (2023, ICCV)~\cite{liu2023aerialvln} & 128.2m & 13.8\% & 35.1\% & 12.0\% & - & 18.3 & - & - & -\\
NaVid (2024, RSS)~\cite{zhang2024navid} & 140.1m & 21.9\% & 38.2\% & 19.9\% & 6.7B & 18.1 & 7.2e8 & 8.6s & 15.6~GB\\
NavMorph (2025, ICCV)~\cite{yao2025navmorph} & 133.7m & 17.8\% & 19.1\% & 15.0\% & - & 11.3 & - & - & -\\
NaVILA (2025, RSS)~\cite{cheng2025navila} & 101.9m & 23.0\% & 50.8\% & 18.8\% & 8B & 23.4 & 3.8e9 & 22.9s & 17.1~GB\\
OpenFly (2026, ICLR)~\cite{gao2026openfly} & 122.8m & 29.1\% & \textbf{55.3\%} & 22.3\% & 6.7B & 20.9 & 10.4e8 & 11.0s & 15.6~GB\\
ARC-VLN (Ours) & \textbf{70.5m} & \textbf{33.4\%} & 44.5\% & \textbf{31.1\%} & \textbf{2.7B} & \textbf{10.7} & \textbf{4.4e8} & \textbf{5.3s} & \textbf{7.1~GB}\\
\midrule
\multicolumn{10}{c}{Test-Unseen}\\
\midrule
Method & NE\ensuremath{\downarrow} & SR\ensuremath{\uparrow} & OSR\ensuremath{\uparrow} & SPL\ensuremath{\uparrow} & Par\ensuremath{\downarrow} & AS\ensuremath{\downarrow} & GFL\ensuremath{\downarrow} & IL\ensuremath{\downarrow} & AIM\ensuremath{\downarrow}\\
\midrule
AerialVLN (2023, ICCV)~\cite{liu2023aerialvln} & 201.1m & 2.7\% & 21.2\% & 2.1\% & - & 21.7 & - & - & -\\
NaVid (2024, RSS)~\cite{zhang2024navid} & 193.7m & 15.8\% & 27.1\% & 10.6\% & 6.7B & 17.9 & 2.1e8 & 8.6s & 15.6~GB\\
NavMorph (2025, ICCV)~\cite{yao2025navmorph} & 105.6m & 22.3\% & 25.3\% & 18.4\% & - & 12.3 & - & - & -\\
NaVILA (2025, RSS)~\cite{cheng2025navila} & 99.8m & 23.3\% & 58.0\% & 17.9\% & 8B & 22.1 & 1.0e9 & 25.4s & 17.1~GB\\
OpenFly (2026, ICLR)~\cite{gao2026openfly} & 158.6m & 27.1\% & \textbf{62.3\%} & 14.1\% & 6.7B & 23.3 & 3.1e8 & 11.7s & 15.6~GB\\
ARC-VLN (Ours) & \textbf{55.7m} & \textbf{29.3\%} & 39.4\% & \textbf{27.3\%} & \textbf{2.7B} & \textbf{10.1} & \textbf{7.1e7} & \textbf{4.2s} & \textbf{7.1~GB}\\
\bottomrule
\end{tabular*}
\endgroup
\end{table*}

\begin{table}[!t]
\centering
\begin{minipage}[t]{\columnwidth}
\centering
\includegraphics[width=\linewidth]{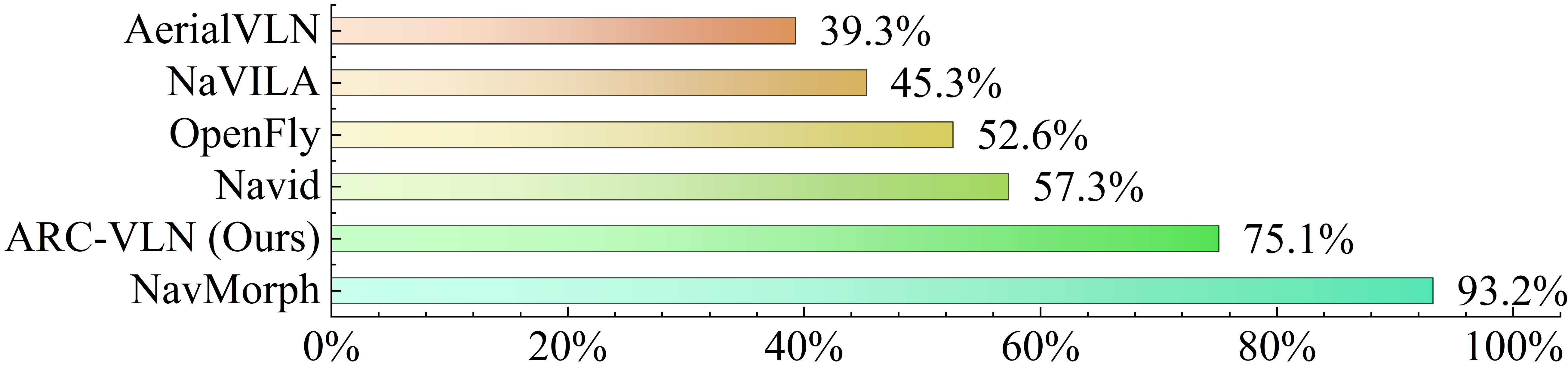}
\captionof{figure}{Stop success conversion rates of ARC-VLN and baseline methods on Test-Seen.}
\label{fig:2}
\end{minipage}
\par\vspace{8pt}
\begin{minipage}[t]{\columnwidth}
\centering
\includegraphics[width=\linewidth]{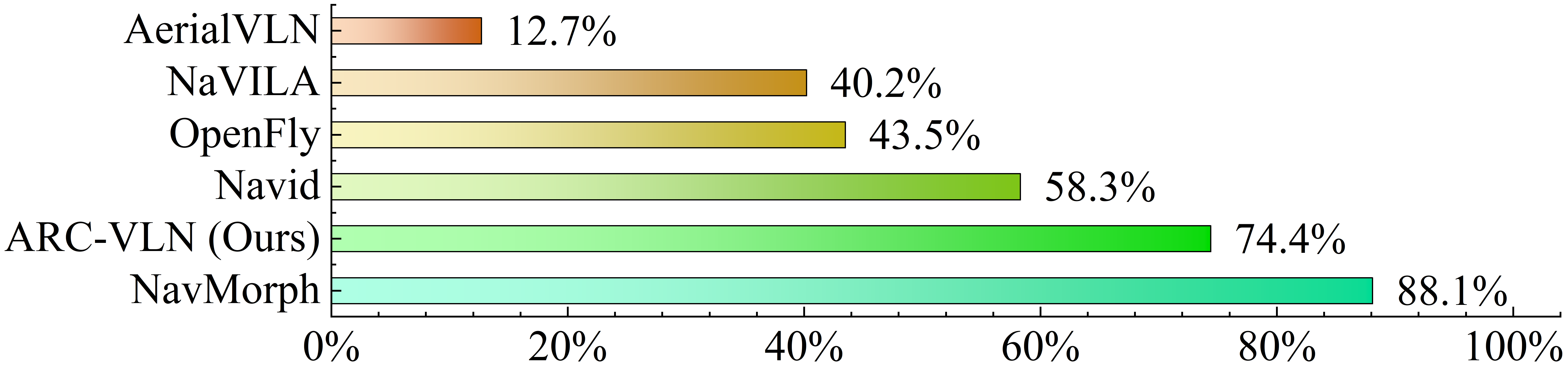}
\captionof{figure}{Stop success conversion rates of ARC-VLN and baseline methods on Test-Unseen.}
\label{fig:3}
\end{minipage}
\end{table}

\Needspace{5\baselineskip}

\subsection{Experimental results}

\subsubsection{Comparison with baseline methods}

Table~\ref{tab:1} compares the navigation performance, parameter efficiency and computational efficiency of ARC-VLN and representative methods on Test-Seen and Test-Unseen. ARC-VLN achieved the lowest NE and highest SR and SPL among the compared methods, with fewer unique parameters and lower inference costs.

\textbf{Task performance.} On Test-Seen, ARC-VLN achieved the best NE (70.5 m), SR (33.4\%) and SPL (31.1\%) among the compared methods. Compared with OpenFly, NE decreased by 42.6\%, with relative SR and SPL gains of 14.8\% and 39.5\%, respectively. NE was also 30.8\% lower than that of NaVILA, which ranked second in NE.

On Test-Unseen, ARC-VLN again achieved the best NE (55.7 m), SR (29.3\%) and SPL (27.3\%). Compared with OpenFly, NE decreased by 64.9\%, with relative SR and SPL gains of 8.1\% and 93.6\%, respectively. Compared with the respective runners-up, NE decreased by 44.2\% relative to NaVILA, while SPL showed a relative gain of 48.4\% over NavMorph. These results support improved navigation accuracy and path efficiency in the evaluated unseen scene.

Notably, ARC-VLN had a lower OSR than OpenFly. To further analyze the model’s ability to convert reaching the goal region into final success, we introduce the Stop Success Conversion Rate (SSCR) as an auxiliary metric. This auxiliary metric measures the model’s ability to translate having reached the goal region at some point into successfully terminating within it. We define SSCR as the ratio of SR to OSR:

\begin{equation}
\mathrm{SSCR}=\frac{\mathrm{SR}}{\mathrm{OSR}}\times100\%.
\label{eq:123}
\end{equation}

OSR records entry into the goal region, whereas SR records successful termination within it. Higher SSCR indicates that a larger proportion of goal-reaching episodes terminate successfully.

As shown in Figs.~\ref{fig:2} and~\ref{fig:3}, ARC-VLN achieved SSCRs of 75.1\% and 74.4\% on Test-Seen and Test-Unseen, respectively, compared with 52.6\% and 43.5\% for OpenFly. These results show that ARC-VLN clearly outperformed OpenFly on this metric. Meanwhile, OpenFly’s AS values were 20.9 and 23.3, respectively, far higher than ARC-VLN’s 10.7 and 10.1. Together, these results indicate that OpenFly’s longer interaction sequences may have provided more opportunities to reach the goal region through additional exploration, partly explaining its higher OSR. However, its lower SSCR indicates that some episodes entered the goal region but subsequently left it and terminated unsuccessfully. ARC-VLN achieved higher SR and SSCR with fewer interaction steps. This pattern is consistent with less redundant movement after reaching the goal region and greater agreement between goal attainment and the final termination state.

NavMorph achieved SSCR values of 93.2\% and 88.1\% on Test-Seen and Test-Unseen, respectively, exceeding those of ARC-VLN. However, its SR values were only 17.8\% and 22.3\%, and its OSR values were only 19.1\% and 25.3\%. This indicates that the high SSCR of NavMorph is based on the fact that fewer trajectories can reach the target neighborhood. Overall, ARC-VLN achieved lower NE and higher SR and SPL while maintaining a high SSCR. This indicates more balanced performance in goal localization, final success, path efficiency and determining where to stop.

\textbf{Parameter efficiency. }ARC-VLN uses 2.7B unique LLM backbone parameters, 59.7\% fewer than NaVid and OpenFly (6.7B) and 66.3\% fewer than NaVILA (8B). As shown in Table~\ref{tab:1}, ARC-VLN achieved lower NE and higher SR and SPL with fewer unique LLM backbone parameters. These results indicate simultaneous improvements in parameter efficiency and navigation performance.

\textbf{Computational efficiency. }On Test-Seen, ARC-VLN reduced AS from 20.9 to 10.7 (a 48.8\% reduction) and GFL by 57.7\% compared with OpenFly. IL decreased from 11.0 s to 5.3 s, a reduction of 51.8\%. On Test-Unseen, AS decreased from 23.3 to 10.1 (a 56.7\% reduction), and GFL decreased by 77.1\%. IL decreased from 11.7 s to 4.2 s, a reduction of 64.1\%. These results show that ARC-VLN incurred lower computational costs and shorter model inference times than OpenFly in both seen and unseen scenes.

The reduction in computational cost primarily arises from two factors. First, token-type-aware routing and active-token compaction reduce per-step computation by excluding early-exiting visual tokens from subsequent recursive blocks. Fewer interaction steps further reduce cumulative inference costs across navigation trajectories.

ARC-VLN reduced AIM by 54.5\% relative to OpenFly on both splits. Fewer unique parameters and active visual tokens contribute to the lower memory demand. Together, lower GFL, IL and AIM improve deployment feasibility on resource-constrained platforms.

\subsubsection{Ablation experiments}

\begin{table}[!t]
\caption{Model configurations for the ablation study of ARC-VLN}
\label{tab:2}
\centering
\begingroup
\footnotesize
\setlength{\tabcolsep}{3pt}
\renewcommand{\arraystretch}{1.12}
\begin{tabular}{@{}lcccc@{}}
\toprule
Variant & PS & DR & $\mathcal{L}_{\mathrm{ent}}$ & Route\\
\midrule
OpenFly & \textbf{\ensuremath{\times}} & \textbf{\ensuremath{\times}} & \textbf{–} & Full Depth\\
Shared Backbone & \textbf{\ensuremath{\checkmark}} & \textbf{\ensuremath{\times}} & \textbf{–} & Full Depth\\
\multirow{3}{*}{w/o $\mathcal{L}_{\mathrm{ent}}$} & \multirow{3}{*}{\textbf{\ensuremath{\checkmark}}} & \multirow{3}{*}{\textbf{\ensuremath{\checkmark}}} & \multirow{3}{*}{\textbf{\ensuremath{\times}}} & Exit\_B0\\
 &  &  &  & Exit\_B1\\
 &  &  &  & Exit\_B2\\
\textbf{ARC}\textbf{-VLN}\textbf{ (Ours)} & \textbf{\ensuremath{\checkmark}} & \textbf{\ensuremath{\checkmark}} & \textbf{\ensuremath{\checkmark}} & Adaptive Depth\\
\bottomrule
\end{tabular}
\endgroup
\par\vspace{8pt}
\begin{minipage}[t]{\columnwidth}
\centering
\includegraphics[width=\linewidth]{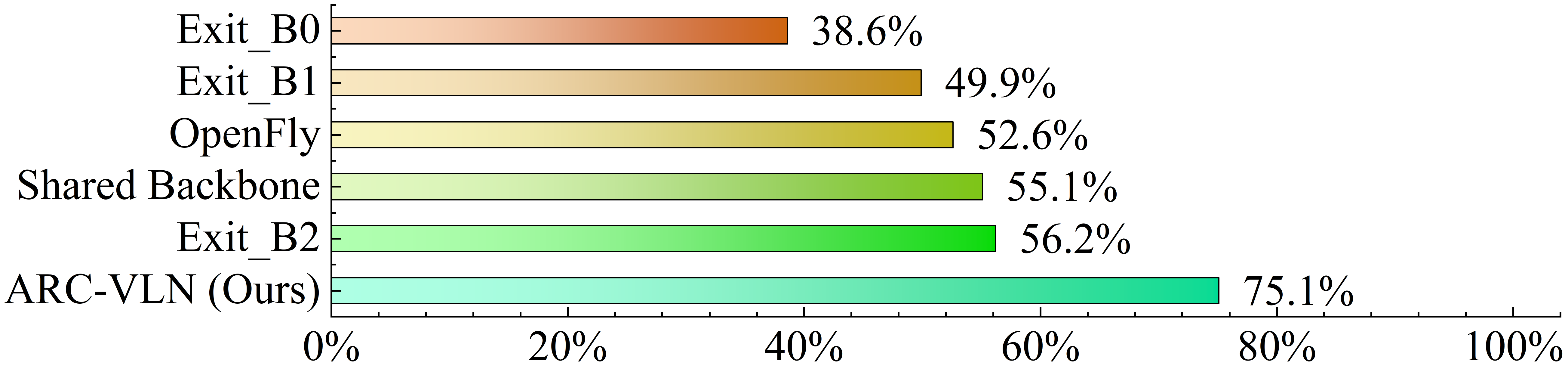}
\captionof{figure}{Stop success conversion rates in the ablation study on Test-Seen.}
\label{fig:4}
\end{minipage}
\par\vspace{8pt}
\begin{minipage}[t]{\columnwidth}
\centering
\includegraphics[width=\linewidth]{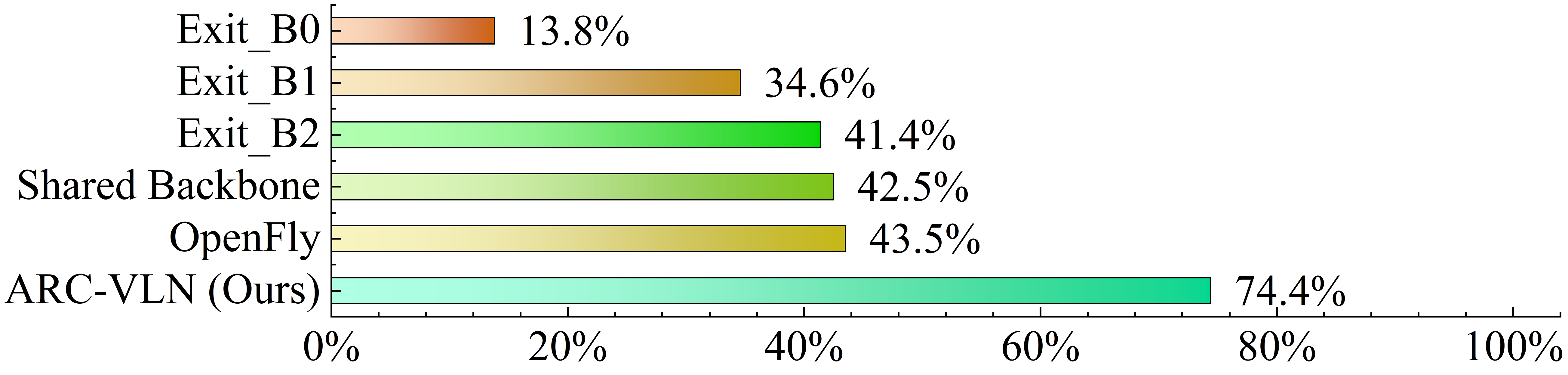}
\captionof{figure}{Stop success conversion rates in the ablation study on Test-Unseen.}
\label{fig:5}
\end{minipage}
\end{table}

\begin{table*}[!t]
\caption{Ablation results for navigation performance and efficiency on the OpenFly benchmark}
\label{tab:3}
\centering
\footnotesize
\setlength{\tabcolsep}{3pt}
\renewcommand{\arraystretch}{1.12}
\begin{tabular*}{\textwidth}{@{\hspace{6pt}\extracolsep{\fill}}*{11}{c}@{\hspace{6pt}}}
\toprule
\multicolumn{11}{c}{Test-Seen}\\
\midrule
\multicolumn{2}{@{\hspace{6pt}}l}{Variant} & NE\ensuremath{\downarrow} & SR\ensuremath{\uparrow} & OSR\ensuremath{\uparrow} & SPL\ensuremath{\uparrow} & Par\ensuremath{\downarrow} & AS\ensuremath{\downarrow} & GFL\ensuremath{\downarrow} & IL\ensuremath{\downarrow} & AIM\ensuremath{\downarrow}\\
\midrule
\multicolumn{2}{@{\hspace{6pt}}l}{OpenFly} & 122.8m & 29.1\% & \textbf{55.3\%} & 22.3\% & 6.7B & 20.9 & 10.4e8 & 11.0s & 15.6~GB\\
\multicolumn{2}{@{\hspace{6pt}}l}{\shortstack[l]{Shared Backbone}} & 101.8m & 27.7\% & 50.3\% & 25.3\% & \textbf{2.7B} & 13.7 & 7.9e8 & 10.7s & \textbf{7.0~GB}\\
\multirow{3}{*}{w/o $\mathcal{L}_{\mathrm{ent}}$} & Exit\_B0 & 146.7m & 18.8\% & 48.7\% & 16.6\% & \textbf{2.7B} & 28.5 & 9.9e8 & 20.6s & 7.1~GB\\
 & Exit\_B1 & 143.5m & 26.0\% & 52.1\% & 23.6\% & \textbf{2.7B} & 20.1 & 9.5e8 & 19.1s & 7.1~GB\\
 & Exit\_B2 & 109.9m & 27.5\% & 48.9\% & 24.7\% & \textbf{2.7B} & 11.1 & 8.4e8 & 11.8s & 7.1~GB\\
\multicolumn{2}{@{\hspace{6pt}}l}{\shortstack[l]{ARC-VLN(Ours)}} & \textbf{70.5m} & \textbf{33.4\%} & 44.5\% & \textbf{31.1\%} & \textbf{2.7B} & \textbf{10.7} & \textbf{4.4e8} & \textbf{5.3s} & 7.1~GB\\
\midrule
\multicolumn{11}{c}{Test-Unseen}\\
\midrule
\multicolumn{2}{@{\hspace{6pt}}l}{Variant} & NE\ensuremath{\downarrow} & SR\ensuremath{\uparrow} & OSR\ensuremath{\uparrow} & SPL\ensuremath{\uparrow} & Par\ensuremath{\downarrow} & AS\ensuremath{\downarrow} & GFL\ensuremath{\downarrow} & IL\ensuremath{\downarrow} & AIM\ensuremath{\downarrow}\\
\midrule
\multicolumn{2}{@{\hspace{6pt}}l}{OpenFly} & 158.6m & 27.1\% & \textbf{62.3\%} & 14.1\% & 6.7B & 23.3 & 3.1e8 & 11.7s & 15.6~GB\\
\multicolumn{2}{@{\hspace{6pt}}l}{\shortstack[l]{Shared Backbone}} & 102.9m & 22.5\% & 52.9\% & 20.0\% & \textbf{2.7B} & 13.3 & 2.0e8 & 17.8s & \textbf{7.0~GB}\\
\multirow{3}{*}{w/o $\mathcal{L}_{\mathrm{ent}}$} & Exit\_B0 & 202.8m & 8.4\% & 60.8\% & 6.9\% & \textbf{2.7B} & 38.2 & 3.3e8 & 23.2s & 7.1~GB\\
 & Exit\_B1 & 154.1m & 20.6\% & 59.5\% & 18.6\% & \textbf{2.7B} & 20.4 & 2.5e8 & 14.7s & 7.1~GB\\
 & Exit\_B2 & 113.2m & 23.0\% & 55.5\% & 20.2\% & \textbf{2.7B} & 14.5 & 2.2e8 & 16.6s & 7.1~GB\\
\multicolumn{2}{@{\hspace{6pt}}l}{\shortstack[l]{ARC-VLN(Ours)}} & \textbf{55.7m} & \textbf{29.3\%} & 39.4\% & \textbf{27.3\%} & \textbf{2.7B} & \textbf{10.1} & \textbf{7.1e7} & \textbf{4.2s} & 7.1~GB\\
\bottomrule
\end{tabular*}
\end{table*}

Table~\ref{tab:2} compares configurations of parameter sharing (PS), dynamic routing (DR) and auxiliary routing losses; \mbox{“–”} marks an absent module. OpenFly is unchanged; Shared Backbone adds PS while retaining full-depth processing for all tokens. The w/o $\mathcal{L}_{\mathrm{ent}}$ configurations retain PS and the routing module but omit $\mathcal{L}_{\mathrm{ent}}$. Spontaneous routing collapse without $\mathcal{L}_{\mathrm{ent}}$
varied in both onset time and exit depth across runs.
Directly evaluating the resulting checkpoints would reflect both
their final exit depths and their differing training histories,
complicating performance comparisons across exit depths.
To obtain controlled comparisons across prescribed exit depths,
we therefore fixed all visual tokens to exit after B0, B1, or B2
throughout training and evaluation, yielding Exit\_B0, Exit\_B1,
and Exit\_B2, respectively. ARC-VLN combines PS, DR and all three losses for adaptive visual-token depths.

\textbf{Effect of parameter sharing. }As shown in Table~\ref{tab:3}, Shared Backbone reduced unique LLM backbone parameters from 6.7B to 2.7B (59.7\%) compared with OpenFly. AIM decreased from 15.6 to 7.0 GB, a reduction of 55.1\%. On Test-Seen and Test-Unseen, respectively, NE decreased by 17.1\% and 35.1\%, with relative SPL gains of 13.5\% and 41.8\%. AS decreased by 34.4\% and 42.9\%, and GFL by 24.0\% and 35.5\%, respectively. On Test-Seen, IL also decreased by 2.7\% relative to OpenFly. However, relative SR reductions were 4.8\% and 17.0\%, and relative OSR reductions were 9.0\% and 15.1\% on Test-Seen and Test-Unseen, respectively. IL on Test-Unseen also increased by 52.1\% relative to OpenFly. Thus, parameter sharing reduced resource demands and improved NE and SPL, but involved trade-offs in success rates and inference latency.

\textbf{Effect of routing collapse. }After $\mathcal{L}_{\mathrm{ent}}$ was removed, routing assignments for visual tokens collapsed to a single exit path. The model consequently lost the ability to assign different computational depths to individual visual tokens. Across the three configurations exhibiting routing collapse, NE, SR and SPL improved with increasing exit depth, although the gains differed markedly between stages. Moving from Exit\_B0 to Exit\_B1 increased SR and SPL by 38.3\% and 42.2\% on Test-Seen, and by 145.2\% and 169.6\% on Test-Unseen, respectively. From Exit\_B1 to Exit\_B2, the corresponding gains fell to 5.8\% and 4.7\% on Test-Seen, and 11.7\% and 8.6\% on Test-Unseen, respectively. Although each transition added one recursive block, the second produced markedly smaller relative gains in SR and SPL. This indicates that performance gains did not scale linearly with computational depth. Compared with Exit\_B1, Exit\_B2 achieved lower NE, AS and GFL on both test splits. However, OSR decreased by 6.1\% and 6.7\% on Test-Seen and Test-Unseen, respectively, and IL increased by 12.9\% on Test-Unseen. Thus, uniformly increasing computational depth can improve some navigation metrics, but the benefits vary across metrics and may be accompanied by deterioration in others.

Exit\_B2 performed relatively well among the three configurations exhibiting routing collapse. Compared with OpenFly, it still achieved better results on some navigation metrics despite having 59.7\% fewer unique parameters and 54.5\% lower AIM. On Test-Seen and Test-Unseen, NE decreased by 10.5\% and 28.6\%, respectively, and SPL increased by 10.8\% and 43.3\%. Both AS and GFL also decreased on the two test splits. However, Exit\_B2 had lower SR and OSR and higher IL than OpenFly. These results show that fewer unique parameters can still support better performance on some navigation metrics. A larger parameter count did not confer consistent advantages across all metrics. Increasing parameter counts or computational depth alone therefore does not guarantee simultaneous improvements across all metrics.

\textbf{Role of cross-entropy regularization and adaptive routing. }At the same unique parameter count, the full ARC-VLN model outperformed Shared Backbone and the three w/o $\mathcal{L}_{\mathrm{ent}}$ configurations on both test splits. Compared with Shared Backbone, ARC-VLN reduced NE by 30.7\% and 45.9\% on Test-Seen and Test-Unseen, respectively. SR increased by 20.6\% and 30.2\%, and SPL increased by 22.9\% and 36.5\%, respectively. GFL decreased by 44.3\% and 64.5\%, and IL decreased by 50.5\% and 76.4\%, respectively. These results indicate that adding DR and its auxiliary losses to a backbone with PS can improve navigation performance while reducing cumulative inference costs per navigation task.

Compared with Exit\_B2, ARC-VLN achieved relative SR increases of 21.5\% and 27.4\% on Test-Seen and Test-Unseen, respectively, with the same unique parameter count and AIM. GFL decreased by 47.6\% and 67.7\% on the two test splits, respectively. Together with the collapse to a single exit depth observed after removing $\mathcal{L}_{\mathrm{ent}}$, these results indicate that cross-entropy regularization helps maintain effective routing assignments across multiple depths. This allows adaptive routing to assign different computational paths to individual visual tokens, yielding better navigation performance and computational efficiency than configurations with a uniform exit depth.

As shown in Figs.~\ref{fig:4} and~\ref{fig:5}, ARC-VLN achieved SSCRs of 75.1\% and 74.4\% on Test-Seen and Test-Unseen, respectively, despite having a lower OSR than the other ablation configurations. Both values exceeded those of Shared Backbone and the three configurations exhibiting routing collapse. This indicates that, among trajectories that entered the goal region, a higher proportion ultimately terminated successfully with ARC-VLN. Together with higher SR and SPL and lower AS, these results show the full model’s advantages in both overall navigation performance and successful termination.

\subsubsection{Routing behaviour analysis}

\begin{figure}[!t]
\centering
\includegraphics[width=\columnwidth]{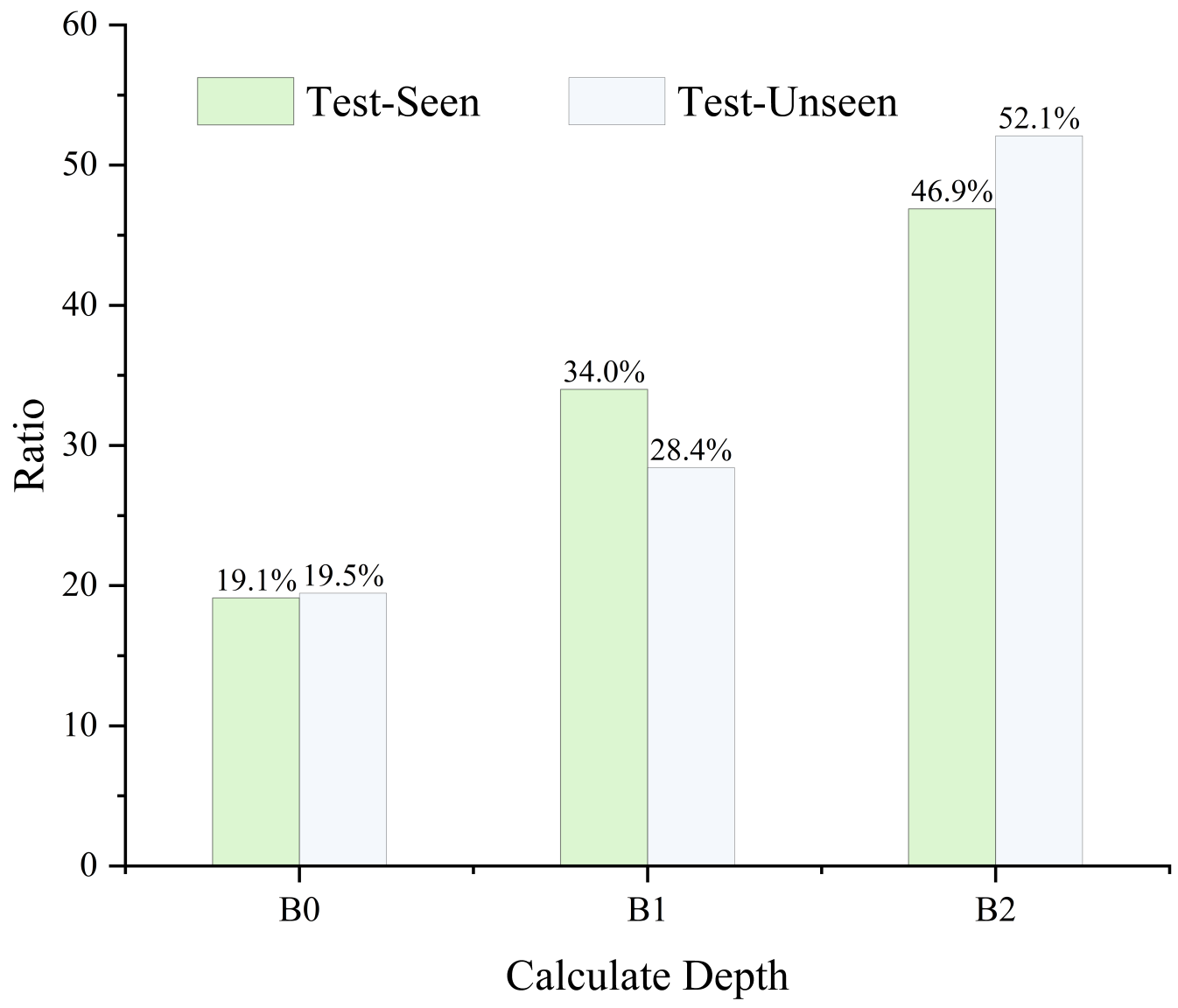}
\caption{Distribution of visual token exit depths in ARC-VLN on Test-Seen and Test-Unseen.}
\label{fig:6}
\end{figure}

\begin{figure}[!t]
\centering
\includegraphics[width=\columnwidth]{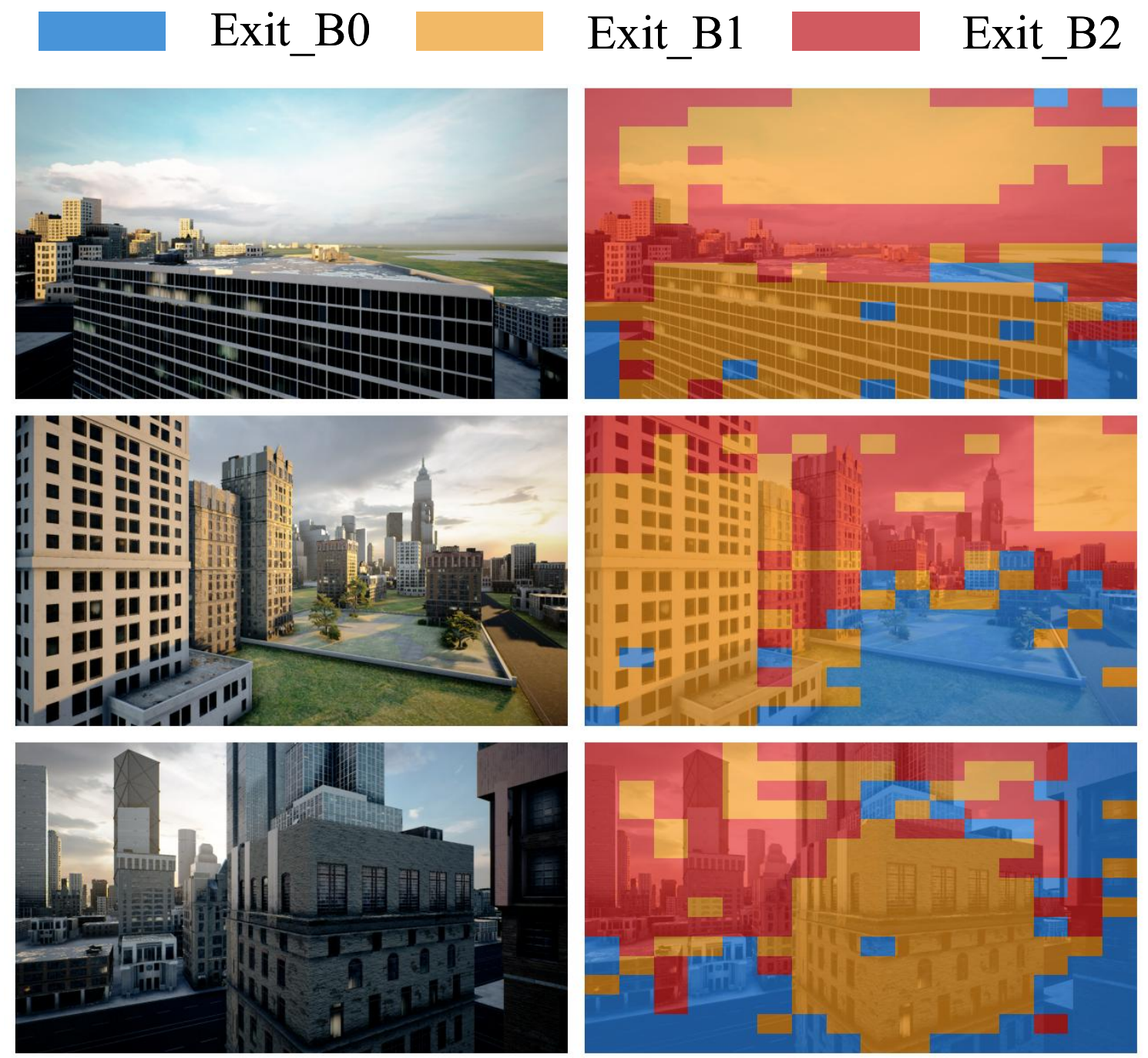}
\caption{Spatial visualization of visual token exit depths in ARC-VLN.}
\label{fig:7}
\end{figure}

We quantified visual-token exit distributions and used spatial visualizations to examine computation allocation in ARC-VLN. B0, B1 and B2 denote exits after 1, 2 and 3 recursive blocks, respectively. Text and action tokens always undergo full recursive computation and were excluded from these statistics.

As shown in Fig.~\ref{fig:6}, visual-token exit proportions at B0, B1 and B2 were 19.1\%, 34.0\% and 46.9\% on Test-Seen, respectively; 53.1\% exited early. The corresponding proportions on Test-Unseen were 19.5\%, 28.4\% and 52.1\%, respectively; 47.9\% exited early. Active-token compaction excludes early-exiting visual tokens from subsequent recursive blocks, shortening sequences and reducing computation.

Fig.~\ref{fig:7} visualizes the spatial distribution of visual-token exit depths, with blue, orange and red denoting B0, B1 and B2, respectively. Different positions within the same observation receive different recursive depths, illustrating adaptive computation allocation at the visual-token level.

\section{Conclusion and future work}

We present ARC-VLN, an aerial VLN framework that combines recursive parameter sharing with token-type-aware adaptive computation to reduce model storage and inference costs.
ARC-VLN shares language-backbone parameters across recursive stages and adaptively assigns exit depths to visual tokens while maintaining full-depth processing for text and action tokens.
Active-token compaction reduces computation in subsequent recursive blocks, while cross-entropy regularization mitigates routing collapse and encourages visual-token allocation across multiple exit depths.
On the evaluated OpenFly Test-Seen and Test-Unseen subsets, ARC-VLN achieves the lowest NE and the highest SR and SPL among the compared methods.
Compared with OpenFly, ARC-VLN reduces the number of unique language-backbone parameters from 6.7B to 2.7B, a reduction of 59.7\%.
It also reduces the average number of decision steps per trajectory, cumulative computational cost, model inference time per trajectory, and GPU memory usage.
These results demonstrate the effectiveness of token-type-aware adaptive recursive computation in improving navigation performance and inference efficiency in aerial VLN.

In the future, we will explore adaptive reasoning strategies that allocate reasoning effort according to navigation difficulty. We will also conduct real-world flight experiments on resource-constrained aerial platforms to evaluate ARC-VLN’s navigation performance, end-to-end latency, and energy consumption.


\IEEEtriggeratref{18}

\end{document}